\documentclass[letterpaper, 10 pt, conference]{ieeeconf}
\IEEEoverridecommandlockouts  
\let\labelindent\relax
\usepackage{enumitem}

\usepackage{graphicx}      
\usepackage{mathtools}
\usepackage{graphics} 
\usepackage[normalem]{ulem}
\usepackage{amsmath,amssymb}
\usepackage{mathrsfs}

\usepackage{graphicx}
\usepackage{varioref}
\usepackage{import}
\usepackage{framed,xcolor}
\usepackage{color}
\usepackage{comment}
\usepackage{multirow}
\usepackage{float}
\usepackage{mathtools}
\usepackage{booktabs}
\usepackage{soul}
\usepackage{amsbsy}
\usepackage{microtype}
\usepackage{comment}
\newtheorem{theorem}{Theorem}

\newtheorem{Definition}{Definition}

\newcommand{\vect}[1]{\ensuremath{\boldsymbol{\mathrm{#1}}}}
\newcommand{\tr}{\operatorname{tr}}
\usepackage{graphicx}
\usepackage{tabularx,booktabs}
\usepackage{color}
\usepackage{multicol}
\usepackage{amsmath,amssymb}
\usepackage{amsfonts}
\usepackage{textcomp}
\usepackage{psfrag}
\usepackage{multimedia}
\usepackage{fancybox}
\usepackage{comment}
\usepackage{soul}

\usepackage[ruled,vlined]{algorithm2e}

\title{\LARGE \bf  Quasi-Newton Compatible Actor--Critic for Deterministic Policies} 
\author{Arash Bahari Kordabad, Dean Brandner, Sebastien Gros, Sergio Lucia,  and Sadegh Soudjani
\thanks{ Arash Bahari Kordabad and Sadegh Soudjani are with the Max Planck Institute for Software Systems, Kaiserslautern, Germany (corresponding e-mail: {\tt\small{arashbk@mpi-sws.org}}).
Sadegh Soudjani is also with the University of Birmingham, Birmingham, United Kingdom.
Sebastien Gros is with the Norwegian University of Science and Technology, Trondheim, Norway. Dean Brandner and Sergio Lucia are with the TU Dortmund, Dortmund, Germany.
}
}

\begin{document}
\bstctlcite{IEEEexample:BSTcontrol}
\maketitle
\thispagestyle{empty}
\pagestyle{empty}

\begin{abstract}
    In this paper, we propose a second-order deterministic actor--critic framework in reinforcement learning that extends the classical deterministic policy gradient method to exploit curvature information of the performance function. Building on the concept of compatible function approximation for the critic, we introduce a quadratic critic that simultaneously preserves the true policy gradient and an approximation of the performance Hessian. A least-squares temporal difference learning scheme is then developed to estimate the quadratic critic parameters efficiently. This construction enables a quasi-Newton actor update using information learned by the critic, yielding faster convergence compared to first-order methods. The proposed approach is general and applicable to any differentiable policy class. Numerical examples demonstrate that the method achieves improved convergence and performance over standard deterministic actor--critic baselines.
\end{abstract}

\vspace{-0.2cm}

\section{Introduction}
Policy gradient methods have become a class of widely used reinforcement learning (RL) algorithms due to their ability to directly optimize parameterized policies through gradient information~\cite{SuttonPG,sutton2018reinforcement}. Among these, deterministic policy gradient (DPG) methods offer a significant advantage in sample efficiency by optimizing policies that are  deterministic mappings from state space to action space instead of optimizing policies in measure spaces that assign probabilities to each possible action in a given state~\cite{silver2014deterministic}. This reduction in sampling variance makes DPG particularly attractive in control-oriented applications where data collection is costly, the dynamics are unknown or partially known, and the optimal control feedback is deterministic~\cite{kordabad2021mpc, cai2021mpc,cai2023energy}.

Most policy gradient methods rely on first-order gradient descent (or ascent) iterations, where the policy parameters are updated solely using the estimated gradient of the performance function. While this approach is effective, it can suffer from slow convergence and sensitivity to learning-rate selection. Second-order policy optimization methods, such as quasi-Newton approaches, accelerate convergence by incorporating curvature information---captured by the Hessian matrix of the performance function---into the update rule~\cite{brandner2025computationally, jha2020quasi, kakade2001natural}. Although exact computation of the Hessian is often impractical in RL, recent developments have proposed tractable Hessian approximations that rely solely on the second derivative of the action-value function with respect to the action and the first derivative of the policy with respect to its parameters~\cite{kordabad2022quasi}. These approximations provide curvature information that is relatively easy to estimate while maintaining low computational cost, leading to faster learning compared with first-order methods.

Actor--critic methods combine the advantages of value-based and policy-based RL by maintaining two separate components: an actor representing the policy and a critic that evaluates the policy~\cite{konda1999actor}. In these methods, the action-value function is approximated by a critic model with parameters that are independent of the underlying policy, and it is trained to match the action-value function associated with the policy. The actor then uses the gradient of the critic to update the policy parameters~\cite{grondman2012survey}. However, when substituting an approximate action-value function into the expressions for the policy gradient and the approximated Hessian, care must be taken to ensure \textit{compatibility} between the critic and the policy~\cite{silver2014deterministic}. Compatibility guarantees that the critic introduces no bias into the gradient or curvature estimates, preserving the true update directions required for convergence.

\textbf{Contributions.} The main contributions of this paper are as follows: i) We propose a compatible quadratic critic that simultaneously preserves the deterministic policy gradient and the approximated Hessian of the performance function. This critic extends the standard linear compatible function approximation by introducing a quadratic term in the action, enabling accurate representation of both gradient and curvature information. ii) We develop a least-squares temporal-difference learning algorithm that efficiently estimates the critic parameters.

\textbf{Outline.} The rest of the paper is organized as follows. Section~\ref{sec:pre} presents the preliminaries, including the Markov decision process setting, the deterministic policy gradient theorem, and the approximated Hessian formulation. Section~\ref{sec:QN_AC} introduces the compatible quadratic critic and the quasi-Newton actor--critic algorithm. Section~\ref{sec:Sim} provides simulation results that illustrate the performance and convergence benefits of the proposed approach. Finally, Section~\ref{sec:con} concludes the paper and discusses directions for future research.
\vspace{-0.01cm}
\section{Preliminaries and Background} \label{sec:pre}
In this section, we recall the Markov decision process (MDP) formulation, review the deterministic policy gradient theorem, and provide the quasi-Newton extension introduced in~\cite{kordabad2022quasi}.

\subsection{MDP Setting}
We consider an MDP defined by the tuple $(\mathcal{S}, \mathcal{A}, p, \ell, \gamma)$, where $\mathcal{S}\subseteq\mathbb{R}^{n_s}$ and $\mathcal{A}\subseteq\mathbb{R}^{n_a}$ denote the state and action spaces, respectively, and $p(\vect s^{+}\,|\,\vect s,\vect a)$ denotes the transition probability density of the subsequent state $\vect s^+$, given current state $\vect s \in \mathcal{S}$ and action $\vect a \in \mathcal{A}$. Each transition imposes a real-valued stage cost $\ell(\vect s,\vect a)$ and $\gamma\in (0, 1]$ is a discount factor. The initial state $\vect s_0$ is distributed according to $p_1 (\vect s_0)$.

A deterministic policy denoted by $\vect \pi : \mathcal{S} \rightarrow \mathcal{A}$, defines a mapping from state space to action space and specifies how the action $\vect a$ is chosen for each state $\vect s$. We consider a parameterized policy $\vect\pi_{\vect\theta}$ with parameter vector $\vect\theta \in \mathbb{R}^{n_{\theta}}$ and seek an optimal policy that minimizes a performance function by adjusting parameter $\vect\theta$. For a given policy $\vect\pi_{\vect\theta}$, the value function $V^{\vect\pi_{\vect\theta}}: \mathcal{S} \rightarrow\mathbb{R}$ and action-value function $Q^{\vect\pi_{\vect\theta}}: \mathcal{S}\times\mathcal{A} \rightarrow\mathbb{R}$ are defined recursively by the Bellman equations
\begin{align*}
    Q^{\vect\pi_{\vect\theta}} (\vect s,\vect a) &= \ell(\vect s,\vect a)+\gamma\mathbb{E}_{p}\left[V^{\vect\pi_{\vect\theta}} (\vect s^+)|\vect s,\vect a\right], \\
V^{\vect\pi_{\vect\theta}} (\vect s) &= Q^{\vect \pi_{\vect\theta}}(\vect s,\vect\pi_{\vect\theta}(\vect s)).
\end{align*}
The performance function $J: \mathbb{R}^{n_{\theta}}\rightarrow\mathbb{R}$ that we aim to minimize is the expected value of the  discounted infinite-horizon sum of stage costs under the closed-loop dynamics induced by policy $\vect\pi_{\vect\theta}$, and is defined as
\begin{equation}\label{eq:J}
    J(\vect\theta) \!=\!\mathbb{E}_{\vect s_0\sim p_1(\vect s_0),\, p}\!\left[\sum_{k=0}^\infty \gamma^{k} \ell(\vect s_k, \vect a_k)\Big | \vect a_k=\vect \pi_{\vect \theta}(\vect s_k)\!\right].
\end{equation} 
Alternative formulations of the performance objective exist, such as finite-horizon or average-cost formulations~\cite{chen2022learning} and non-Markovian reward machines~\cite{kazemi2020formal,lavaei2020formal,kazemi2022translating}.
However, in this work we focus on the infinite-horizon discounted formulation, which is the most common setting in RL and control. For notational simplicity, the right-hand side of~\eqref{eq:J} can be written as $\mathbb{E}_{\vect s}\left[\ell(\vect s, \vect \pi_{\vect\theta}(\vect s))\right]$, where $\mathbb{E}_{\vect s}[\cdot]$ denotes the expected sum of the discounted state distribution of the MDP in closed-loop with the policy $\vect\pi_{\vect\theta}$.  The policy optimization problem is then formulated as
\begin{align} \label{eq:minJ}
\vect \theta^\star \in \mathrm{arg}\,\min_{\vect \theta} J(\vect \theta).
\end{align}

\subsection{Deterministic Policy Gradient}
The deterministic policy gradient theorem~\cite{silver2014deterministic} provides an explicit expression for the gradient of the performance function $J(\vect\theta)$ with respect to policy parameters:
\begin{equation}\label{eq:PG}
     \nabla_{\vect \theta} J(\vect \theta) =\mathbb{E}_{\vect s} \Big[\nabla_{\vect \theta} \vect \pi_{\vect \theta}(\vect s)\nabla_{\vect a} Q^{\vect \pi_{\vect \theta}}(\vect s,\vect a)\big|_{\vect a=\vect \pi_{\vect \theta}(\vect s)}\Big].
\end{equation}
Intuitively,  $J(\vect\theta)$ is the expectation of the value function $V^{\vect\pi_{\vect\theta}}$ over the initial states; hence,  $\nabla_{\vect \theta} J$ depends on how the value function changes with the policy, which in turn is captured through the gradient of the action--value function $Q^{\vect\pi_{\vect\theta}}$ with respect to the action, evaluated at the action given by the policy. For the exact derivation, see~\cite{silver2014deterministic}.
 More specifically, this result shows that the policy gradient depends only on the gradient of the action-value function with respect to the action and the sensitivity of the policy with respect to its parameters. Most policy gradient algorithms use \eqref{eq:PG} and rely on first-order updates of the form
\begin{equation*}
    \vect \theta_{i+1} = \vect \theta_i - \alpha_{\theta} \nabla_{\vect \theta} J(\vect\theta)|_{\vect \theta=\vect \theta_i},
\end{equation*}
where $\alpha_{\theta}>0$ is a learning rate. The index $i$ denotes the RL algorithm steps which can be different from the actual system runtime index $k$ in general. Although first-order methods are effective, they often suffer from slow convergence, particularly when the optimization landscape is ill-conditioned or when the gradient magnitudes vary across parameters.

\subsection{Quasi-Newton Policy Optimization}
Second-order optimization methods exploit curvature information of the objective function to accelerate convergence. Among these, quasi-Newton methods approximate the Hessian of the objective to adapt the step size and direction in each iteration. The update rule takes the form
\begin{equation}\label{eq:QN:update}
    \vect \theta_{i+1}=\vect \theta_i-\alpha_\theta H^{-1}(\vect \theta_i) \nabla_{\vect \theta} J(\vect \theta)|_{\vect \theta=\vect \theta_i},
\end{equation}
where $H(\vect \theta_i)$ is an approximation of the Hessian of the performance function $J(\vect \theta_i)$, i.e., $H(\vect \theta)\approx \nabla^2_{\vect \theta} J(\vect \theta)$.

Computing the exact Hessian of $J(\vect \theta)$ is intractable in general since it involves second-order derivatives through the recursive dependence of the state distribution and the value function on the policy. To overcome this difficulty, \cite{kordabad2022quasi} proposed an approximated Hessian that depends only on quantities readily available from the policy and the action-value function, given in the following theorem.

\begin{theorem}
For a differentiable deterministic policy $\vect\pi_{\vect\theta}(\vect s)$ and twice continuously differentiable action-value function
$Q^{\vect\pi_{\vect\theta}} (\vect s,\vect a)$, an approximation of the Hessian of $J(\vect \theta)$ can be expressed as
\begin{equation}\label{eq:Grad_Hess_final}
       H(\vect \theta){=}  \mathbb{E}_{\vect s} \Big[ \nabla_{\vect \theta} \vect \pi_{\vect \theta}(\vect s)\nabla_{\vect a}^2 Q^{\vect \pi_{\vect \theta}}(\vect s,\vect a)\Big|_{\vect a=\vect \pi_{\vect \theta}(\vect s)}\!\!\nabla_{\vect \theta} \vect \pi_{\vect \theta}(\vect s)^\top\Big].
\end{equation}
This approximation captures the exact Hessian at the optimal policy parameters, that is, $H(\vect \theta^\star)= \nabla^2_{\vect \theta} J(\vect \theta)|_{\vect \theta=\vect \theta^\star}$, where $\vect \theta^\star$ denotes the optimal solution of the policy optimization problem given in~\eqref{eq:minJ}.
\end{theorem}

\textit{Proof.}    
    See~\cite{kordabad2022quasi} and~\cite{brandner2025computationally}. \hfill $\blacksquare$

The quasi-Newton formulation in~\eqref{eq:Grad_Hess_final} provides a tractable way to incorporate curvature information of the performance function and the approximation becomes exact at the optimal parameters, thereby enabling superlinear convergence rate for the learning under standard regularity conditions.

In practice, both the gradient and the approximated Hessian in~\eqref{eq:Grad_Hess_final} depend only on the action-value function $Q^{\vect\pi_{\vect\theta}}$ and the policy derivatives, which can be computed or estimated efficiently in model-free settings. This makes the approach applicable to a wide range of control problems where direct Hessian computation is otherwise intractable.

The expressions in~\eqref{eq:PG} and~\eqref{eq:Grad_Hess_final} assume access to the exact action-value function of the policy. However, when the true action-value function $Q^{\vect\pi_{\vect\theta}}$ is replaced by an independently parameterized critic $Q^{\vect w}$, the gradient and Hessian approximations in~\eqref{eq:PG} and ~\eqref{eq:Grad_Hess_final} may become biased unless the critic is designed to be \textit{compatible} with the policy (see e.g.,~\cite{qiu2021finite}). The next section formalizes this notion and develops a compatible quadratic critic that enables a consistent quasi-Newton deterministic actor--critic algorithm.

\section{Quasi-Newton Actor--Critic Algorithm}\label{sec:QN_AC}
In this section, we extend the quasi-Newton deterministic policy optimization framework to an actor--critic architecture. The goal is to design an actor update that utilizes curvature information, while allowing the critic to approximate the action-value function independently from the policy parameters. To achieve this, we derive a class of compatible critics whose approximation preserves both the deterministic policy gradient and the quasi-Newton Hessian structure introduced in Section~\ref{sec:pre}. The resulting algorithm integrates the advantages of second-order optimization with the sample efficiency of actor--critic learning.

\subsection{Compatible Quasi-Newton Actor--Critic Framework}

In practice, the true action-value function $Q^{\vect \pi_{\vect \theta}}(\vect s,\vect a)$ in the deterministic policy gradient and quasi-Newton expressions~\eqref{eq:PG}–\eqref{eq:Grad_Hess_final} is not known exactly. Instead, it is approximated by a differentiable parametric function $Q^{\vect w}(\vect s,\vect a)$ with parameters $\vect w$. The critic aims to learn $Q^{\vect w}(\vect s,\vect a)\approx Q^{\vect \pi_{\vect \theta}}(\vect s,\vect a)$, while the actor updates the policy parameters $\vect \theta$ in the gradient or quasi-Newton direction. However, substituting an approximate $Q^{\vect w}$ directly into the deterministic policy gradient or Hessian generally introduces bias, since the derivatives of $Q^{\vect w}$ may not match those of the true $Q^{\vect \pi_{\vect \theta}}$. To ensure that the actor update remains consistent with the true underlying optimization problem, we define next the concept of compatible critic in quasi-Newton update that preserves the exact forms of both the gradient and the approximated Hessian.

\begin{Definition}\textbf{(Quasi-Newton Compatible Critic).}\label{def:comp} A twice continuously differentiable critic $Q^{\vect w}(\vect s,\vect a)$ is \textit{quasi-Newton compatible} with a deterministic policy $\vect\pi_{\vect\theta}(\vect s)$ if there exist parameter values $\vect w$ such that, when $Q^{\vect w}$ is substituted into the actor updates, it preserves both the deterministic policy gradient and the quasi-Newton curvature:
    \begin{subequations}
    \begin{align}
      & \nabla_{\vect \theta} J(\vect \theta) =\mathbb{E}_{\vect s} \Big[\nabla_{\vect \theta} \vect \pi_{\vect \theta}(\vect s)\nabla_{\vect a} Q^{\vect w}(\vect s,\vect a)\big|_{\vect a=\vect \pi_{\vect \theta}(\vect s)}\Big],\label{eq:comp:G}\\
        & H(\vect \theta) {=}  \mathbb{E}_{\vect s} \Big[ \nabla_{\vect \theta} \vect \pi_{\vect \theta}(\vect s)\nabla_{\vect a}^2 Q^{\vect w}(\vect s,\vect a)\Big|_{\vect a=\vect \pi_{\vect \theta}(\vect s)}\nabla_{\vect \theta} \vect \pi_{\vect \theta}(\vect s)^\top\Big].\label{eq:comp:H}
\end{align}
\end{subequations}
\end{Definition}

We now present a theorem that provides sufficient conditions under which a differentiable critic becomes quasi-Newton compatible with the policy.

\begin{theorem}\label{thm:comp:2nd}
If a twice continuously differentiable critic $Q^{\vect w}(\vect s,\vect a)$ satisfies conditions~\ref{cond:i} and \ref{cond:ii} below, then it is a quasi-Newton compatible critic according to Definition~\ref{def:comp}:

\begin{enumerate}[label=(\roman*), left=0pt, itemsep=1.1\baselineskip, parsep=0pt]
    \item \label{cond:i} $\nabla_{\vect a} Q^{\vect w}(\vect s,\vect a)\big|_{\vect a=\vect \pi_{\vect \theta}(\vect s)}=\nabla_{\vect \theta} \vect \pi_{\vect \theta}(\vect s)^\top \vect g$,
    \item \label{cond:ii} $\nabla^2_{\vect a} Q^{\vect w}(\vect s,\vect a)\big|_{\vect a=\vect \pi_{\vect \theta}(\vect s)}=\nabla_{\vect \theta} \vect \pi_{\vect \theta}(\vect s)^\top W \nabla_{\vect \theta} \vect \pi_{\vect \theta}(\vect s)$,
    \end{enumerate} 
  where $\vect g$ minimizes
    $
        \mathbb{E}_{\vect s}[\|\varepsilon_1(\vect s; \vect \theta,\vect g)\|^2],
  $
    with 
    \begin{align*}
    \varepsilon_1(\vect s; \vect \theta,\vect g)&:= \nabla_{\vect \theta} \vect \pi_{\vect \theta}(\vect s)^\top \vect g-\nabla_{\vect a} Q^{\vect \pi_{\vect \theta}}(\vect s,\vect a)\big|_{\vect a=\vect \pi_{\vect \theta}(\vect s)},
    \end{align*}
and $W\succeq 0$ minimizes
    $ \mathbb{E}_{\vect s}[\|\varepsilon_2(\vect s; \vect \theta,W)\|_F^2 ],$
            with $\|\cdot\|_F$ being the Frobenius norm, and
        \begin{align*}
    \!\!\varepsilon_2(\vect s; \!\vect \theta,\!W)\!\!:= \!\!\nabla\!_{\vect \theta} \vect \pi_{\vect \theta}(\vect s)\!^\top\! W \nabla_{\vect \theta} \vect \pi_{\vect \theta}(\vect s)\!-\!\!\nabla^2_{\!\vect a} Q^{\vect \pi_{\vect \theta}}\!(\vect s,\!\vect a)\big|_{\vect a=\!\vect \pi_{\vect \theta}(\vect s)},
    \end{align*}
for parameters $\vect w:=\left\{\vect g\in\mathbb{R}^{n_\theta}, W^{n_\theta\times n_\theta}\right\}$.
\end{theorem}

\textit{Proof. }
    The gradient matching in~\eqref{eq:comp:G} follows from the compatibility result in Theorem~3 in~\cite{silver2014deterministic} using condition~\ref{cond:i} in the theorem. For~\eqref{eq:comp:H}, first we observe,
\begin{align*}
 \bigl\|\varepsilon_2&(\vect s;\vect\theta,W)\bigr\|_F^2
  \\ =& \tr \Big( \big(\nabla_{\vect\theta}\vect\pi_{\vect\theta}(\vect s)^\top W \nabla_{\vect\theta}\vect\pi_{\vect\theta}(\vect s) - \left.\nabla_{\vect a}^2 Q^{\vect \pi_{\vect\theta}}(\vect s,\vect a)\right|_{\vect a=\vect\pi_{\vect\theta}(\vect s)}\big)^\top\\ &\quad \,\,\,\, \big(\nabla_{\vect\theta}\vect\pi_{\vect\theta}(\vect s)^\top W \nabla_{\vect\theta}\vect\pi_{\vect\theta}(\vect s) - \left.\nabla_{\vect a}^2 Q^{\vect \pi_{\vect\theta}}(\vect s,\vect a)\right|_{\vect a=\vect\pi_{\vect\theta}(\vect s)}\big) \Big) \\
  =& \tr \big(W \nabla_{\vect\theta}\vect\pi_{\vect\theta}(\vect s) \nabla_{\vect\theta}\vect\pi_{\vect\theta}(\vect s)^\top W \nabla_{\vect\theta}\vect\pi_{\vect\theta}(\vect s) \nabla_{\vect\theta}\vect\pi_{\vect\theta}(\vect s)^\top\big)\!\\ -2&\tr\big(W \nabla_{\vect\theta}\vect\pi_{\vect\theta}(\vect s) \left.\nabla_{\vect a}^2 Q^{\vect \pi_{\vect\theta}}(\vect s,\vect a)\right|_{\vect a=\vect\pi_{\vect\theta}(\vect s)} \nabla_{\vect\theta}\vect\pi_{\vect\theta}(\vect s)^\top\big)\\ +&\tr\big(\left.\nabla_{\vect a}^2 Q^{\vect \pi_{\vect\theta}}(\vect s,\vect a)\right|_{\vect a=\vect\pi_{\vect\theta}(\vect s)}^\top \left.\nabla_{\vect a}^2 Q^{\vect \pi_{\vect\theta}}(\vect s,\vect a)\right|_{\vect a=\vect\pi_{\vect\theta}(\vect s)}\big).
\end{align*}
Then, using the following matrix calculus identities (see e.g., Equations~(100) and~(114) in \cite{petersen2008matrix}):
\begin{align*}
   & \nabla_W \operatorname{tr}\big(W \nabla_{\vect\theta}\vect\pi_{\vect\theta}(\vect s) \nabla_{\vect\theta}\vect\pi_{\vect\theta}(\vect s)^\top W \nabla_{\vect\theta}\vect\pi_{\vect\theta}(\vect s) \nabla_{\vect\theta}\vect\pi_{\vect\theta}(\vect s)^\top\big) =\\ &\qquad\qquad 2 \nabla_{\vect\theta}\vect\pi_{\vect\theta}(\vect s) \nabla_{\vect\theta}\vect\pi_{\vect\theta}(\vect s)^\top W \nabla_{\vect\theta}\vect\pi_{\vect\theta}(\vect s) \nabla_{\vect\theta}\vect\pi_{\vect\theta}(\vect s)^\top, \\  &\nabla_W \operatorname{tr}\big(W \nabla_{\vect\theta}\vect\pi_{\vect\theta}(\vect s) \left.\nabla_{\vect a}^2 Q^{\vect \pi_{\vect\theta}}(\vect s,\vect a)\right|_{\vect a=\vect\pi_{\vect\theta}(\vect s)} \nabla_{\vect\theta}\vect\pi_{\vect\theta}(\vect s)^\top \big)=\\ &\qquad\quad\quad \nabla_{\vect\theta}\vect\pi_{\vect\theta}(\vect s) \left.\nabla_{\vect a}^2 Q^{\vect \pi_{\vect\theta}}(\vect s,\vect a)\right|_{\vect a=\vect\pi_{\vect\theta}(\vect s)} \nabla_{\vect\theta}\vect\pi_{\vect\theta}(\vect s)^\top, 
\end{align*}
the derivation of $ \bigl\|\varepsilon_2(\vect s;\vect\theta,W)\bigr\|_F^2$ with respect to $W$ becomes
\begin{align}\label{eq:dFdw}
   \nabla_{W} \|\varepsilon_2&(\vect s; \vect \theta,W)\|_F^2 \nonumber \\&= 2 \nabla_{\vect\theta}\vect\pi_{\vect\theta}(\vect s) \nabla_{\vect\theta}\vect\pi_{\vect\theta}(\vect s)^\top W \nabla_{\vect\theta}\vect\pi_{\vect\theta}(\vect s) \nabla_{\vect\theta}\vect\pi_{\vect\theta}(\vect s)^\top\nonumber\\ &\quad- 2\nabla_{\vect\theta}\vect\pi_{\vect\theta}(\vect s) \left.\nabla_{\vect a}^2 Q^{\vect \pi_{\vect\theta}}(\vect s,\vect a)\right|_{\vect a=\vect\pi_{\vect\theta}(\vect s)} \nabla_{\vect\theta}\vect\pi_{\vect\theta}(\vect s)^\top\nonumber \\ &= 2  \nabla_{\vect\theta}\vect\pi_{\vect\theta}(\vect s)\big(\nabla_{\vect\theta}\vect\pi_{\vect\theta}(\vect s)^\top W \nabla_{\vect\theta}\vect\pi_{\vect\theta}(\vect s) \nonumber\\ &\quad -\left.\nabla_{\vect a}^2 Q^{\vect \pi_{\vect\theta}}(\vect s,\vect a)\right|_{\vect a=\vect\pi_{\vect\theta}(\vect s)}\big)\nabla_{\vect\theta}\vect\pi_{\vect\theta}(\vect s)^\top \nonumber \\ &=2  \nabla_{\vect\theta}\vect\pi_{\vect\theta}(\vect s)\varepsilon_2(\vect s; \vect \theta,W)\nabla_{\vect\theta}\vect\pi_{\vect\theta}(\vect s)^\top.
\end{align}
    Using~\eqref{eq:dFdw}, the first-order optimality condition of the objective function $ \mathbb{E}_{\vect s}[\|\varepsilon_2(\vect s; \vect \theta,W)\|_F^2 ]$ yields: 
\begin{align*}
&\nabla_{W} \mathbb{E}_{\vect s}[\|\varepsilon_2(\vect s; \vect \theta,W)\|_F^2]\!=\!0
      \Rightarrow  \mathbb{E}_{\vect s}[\nabla_{W} \|\varepsilon_2(\vect s; \vect \theta,W)\|_F^2]\!=\!0\\ &\Rightarrow\mathbb{E}_{\vect s}\! \Big[ \nabla_{\vect\theta}\vect\pi_{\vect\theta}(\vect s) \varepsilon_2(\vect s; \vect \theta,W)\nabla_{\vect\theta}\vect\pi_{\vect\theta}(\vect s)^\top\Big]\!=\!0\\
     &\Rightarrow\,  \mathbb{E}_{\vect s} \Big[ \nabla_{\vect\theta}\vect\pi_{\vect\theta}(\vect s) \nabla_{\vect a}^2 Q^{\vect w}(\vect s,\vect a)\Big|_{\vect a=\vect \pi_{\vect \theta}(\vect s)}\nabla_{\vect\theta}\vect\pi_{\vect\theta}(\vect s)^\top\Big]{=} \\ &\quad\,\,\,\,  \mathbb{E}_{\vect s} \Big[ \nabla_{\vect\theta}\vect\pi_{\vect\theta}(\vect s) \nabla_{\vect a}^2 Q^{\vect \pi_{\vect\theta}}(\vect s,\vect a)\Big|_{\vect a=\vect \pi_{\vect \theta}(\vect s)} \nabla_{\vect\theta}\vect\pi_{\vect\theta}(\vect s)^\top\Big],
    \end{align*}
    which results in~\eqref{eq:comp:H}. \hfill $\blacksquare$

Intuitively, $W$ describes the local curvature of the action-value function in the neighborhood of the current policy, while $\vect g$ provides the first-order direction of improvement. Note that from condition~\ref{cond:ii}, $W$ is symmetric since $Q^{\vect w}(\vect s,\vect a)$ is twice continuously differentiable.

This theorem generalizes the compatible function approximation of~\cite{silver2014deterministic} to the second-order setting. It is worth noting that this formulation implicitly seeks to satisfy two approximation objectives with a single critic. This requires a sufficiently rich parameterization to capture both properties simultaneously, which will be addressed in the next section.

\subsection{Constructing a Quasi-Newton Compatible Critic}
For any deterministic policy $\vect\pi_{\vect\theta}(\vect s)$, there always exists a quasi-Newton compatible critic of the form
\begin{align*}
    Q^{\vect w}(\vect s,\vect a)=A^{\vect w}(\vect s,\vect a)+V^{\vect v}(\vect s),
\end{align*}
where $V^{\vect v}(\vect s)$ is a baseline function approximating the value function with parameters $\vect v$ that can be in the form of
\begin{equation}\label{eq:baseline}
    V^{\vect v}(\vect s)=\vect v^\top \phi(\vect s),
\end{equation}
with state features $\phi(\vect s)$, and $A^{\vect w}(\vect s,\vect a)$ is the advantage function. With a slight abuse of notation here, the superscript $\vect w$ on $Q^{\vect w}$ collectively denotes the parameters $\{\vect g, \vect W, \vect v\}$, 
while on $A^{\vect w}$ it refers only to the parameters $\{\vect g, \vect W\}$. The advantage function for the second-order learning can take the following form 
\begin{equation}\label{eq:advan}
    A^{\vect w}(\vect s,\vect a)=\psi(\vect s,\vect a)^\top W \psi(\vect s,\vect a)+\psi(\vect s,\vect a)^\top \vect g,
\end{equation}
where $\psi(\vect s,\vect a)= \nabla_{\vect \theta} \vect \pi_{\vect \theta}(\vect s)(\vect a-\vect\pi_{\vect\theta}(\vect s))$ is the state-action feature vector, and $W\succeq 0$ and $\vect g$ are critic parameters associated with curvature and gradient information, respectively. Note that the term $\vect a-\vect\pi_{\vect\theta}(\vect s)$ represents the exploration component in RL. This is typically modeled as a small, isotropic random perturbation. The small isotropic exploration is essential for obtaining unbiased gradient (and Hessian) estimates in theory; see~\cite{kordabad2021bias,gros2021bias} for a detailed discussion of this aspect.

In the first-order deterministic actor--critic method of~\cite{silver2014deterministic}, the advantage function $A^{\vect w}(\vect s,\vect a)$ is modeled as a linear function of the action deviation. Motivated by this, the quasi-Newton compatible critic proposed in this work extends this idea by adding a quadratic term that captures second-order curvature information of the action-value function.

The quadratic structure in~\eqref{eq:advan} ensures that conditions~\ref{cond:i} and \ref{cond:ii} in the theorem are satisfied.

To obtain $\vect g$ and $W\succeq 0$, one would ideally solve two least-squares regression problems to identify the parameters $\vect g$ and $W\succeq 0$ that best match the true action-gradient and action-Hessian of $Q^{\vect \pi_{\vect \theta}}$, i.e.,
\begin{align*}
    \nabla_{\vect a} Q^{\vect w}(\vect s,\vect a)\big|_{\vect a=\vect \pi_{\vect \theta}(\vect s)}&\approx\nabla_{\vect a} Q^{\vect \pi_{\vect\theta}}(\vect s,\vect a)\big|_{\vect a=\vect \pi_{\vect \theta}(\vect s)},\\ \nabla^2_{\vect a} Q^{\vect w}(\vect s,\vect a)\big|_{\vect a=\vect \pi_{\vect \theta}(\vect s)}&\approx\nabla^2_{\vect a} Q^{\vect \pi_{\vect\theta}}(\vect s,\vect a)\big|_{\vect a=\vect \pi_{\vect \theta}(\vect s)}.
\end{align*}
Directly performing this gradient and Hessian matching is, however, challenging in practice because obtaining unbiased samples of $\nabla_{\vect a} Q^{\vect \pi_{\vect\theta}}$ and $\nabla^2_{\vect a} Q^{\vect \pi_{\vect\theta}}$ is generally infeasible. As discussed in~\cite{silver2014deterministic}, even in the first-order case, learning an unbiased estimate of the action–gradient requires access to the true gradient of the action--value function, which is rarely available. In practice, therefore, one resorts to standard policy-evaluation techniques that learn $Q^{\vect w}(\vect s,\vect a)\!\approx\! Q^{\vect \pi_{\vect\theta}}(\vect s,\vect a)$, and smooth function approximators ensure that their gradients and Hessians approximately match as well. Note that value fitting does not necessarily imply accurate gradient and curvature fitting, although in practice the mismatch is often negligible for smooth critics.

\subsection{Second-Order LSTD Algorithm}

A more practical and efficient approach is thus to fit the critic $Q^{\vect w}(\vect s,\vect a)$ directly to the true action--value function $Q^{\vect \pi_{\vect \theta}}(\vect s,\vect a)$ rather than matching their derivatives explicitly. More specifically, the critic parameters $\vect w$ can be obtained by solving a batch least–squares (LS) problem that minimizes the mean-squared error between the true and approximate action--value functions:
\begin{align}
\label{eq:error}
    \min_{\vect w} \mathbb{E}\left[\big(Q^{\vect\pi_{\vect\theta}}(\vect s,\vect a)
      - Q^{\vect w}(\vect s,\vect a)\big)^2\right].
\end{align}

In standard actor--critic algorithms, such critic parameters are updated incrementally using temporal-difference updates at each time step, which can lead to high variance and slow convergence. Rather than updating parameters $\vect v$ and $\vect w=\left\{\vect g,W\right\}$ incrementally, the LS problem~\eqref{eq:error} can be solved in closed form using the \emph{least–squares temporal difference} (LSTD) method. LSTD belongs to the family of batch RL algorithms that estimate the critic parameters from an entire set of collected transitions, offering higher sample efficiency, lower variance, and improved numerical robustness compared with online temporal-difference learning~\cite{cai2023energy}. This closed-form approach is particularly efficient here because $Q^{\vect w}$ is linearly parameterized in the critic weights, allowing the LSTD equations to be derived and solved exactly. To practically identify the critic parameters, the LSTD method decomposes the joint estimation problem in~\eqref{eq:error} into sequential regression steps for the value, gradient, and curvature components, each solved in closed form.

First, the baseline value function parameters $\vect v$ are obtained as follows:
\begin{align}\label{eq:lstdv}
    &\mathbb{E}[\phi(\vect s_k)\bigl(\phi(\vect s_k)\!-\!\gamma\phi(\vect s_{k+1})\bigr)^\top\!] \vect v\!= \!\mathbb{E}[\ell (\vect s_k,\!\vect a_k)\phi(\vect s_k)],
\end{align}
which gives the closed-form solution minimizing the value-based TD error $\mathbb{E}[\delta_k^2]$, where
\begin{equation}\label{eq:delta}
    \delta_k:=\ell(\vect
     s_k,\vect a_k) + \gamma V(\vect s_{k+1}) - V(\vect s_k).
\end{equation}

Next, the gradient parameters $\vect g$ are obtained by matching the linear term of~\eqref{eq:advan} with the TD residual $\delta_k$ according to $\mathbb{E}[(\psi(\vect s_k,\vect a_k)^\top \vect g-\delta_k)^2]$. The optimal solution is achieved by
\begin{equation}\label{eq:ls-g}
     \mathbb{E}[\psi(\vect s_k,\vect a_k)\psi(\vect s_k,\vect a_k)^\top]\, \vect g= \mathbb{E}[\delta_k\psi(\vect s_k,\vect a_k)],
\end{equation}
where $\delta_k$ is evaluated from~\eqref{eq:delta} when the baseline value function in~\eqref{eq:baseline} is used with parameters obtained in~\eqref{eq:lstdv}. Note that~\eqref{eq:ls-g} is aligned with the conventional LSTD actor--critic update (see e.g.,~\cite{sutton2009fast} and~\cite{cai2021mpc}).

Once the gradient component $\vect g$ is determined, the remaining residuals in the TD error are captured through the curvature term, represented by the symmetric matrix $W$ that describes how the action–value function bends locally around the current policy. In order to estimate the curvature matrix $W$, we solve 
\begin{equation}\label{eq:LS:W}
    \min_{W\succeq 0 } \mathbb{E}[(\psi(\vect s_k,\vect a_k)^\top W \psi(\vect s_k,\vect a_k)+\psi(\vect s_k,\vect a_k)^\top \vect g-\delta_k)^2],
\end{equation}  
that matches the quadratic advantage function~\eqref{eq:advan} with the TD residual $\delta_k$ with $\vect g$ obtained from~\eqref{eq:ls-g}. Since the objective in~\eqref{eq:LS:W} is quadratic in $W$ and the constraint $W\succeq 0$ is linear, one can solve~\eqref{eq:LS:W} directly using semidefinite programming (SDP). Another efficient approach is to solve the unconstrained version of~\eqref{eq:LS:W} using the LS technique and then project the solution to the positive semidefinite (PSD) cone.  More specifically, we define:
    \begin{equation*}
\hat{\delta}_k \!\!:= \!\delta_k \!- \psi(\vect s_k,\vect a_k)\!^\top\! \vect g,\,\,
\hat{\psi}(\vect s_k,\vect a_k) \!\!:=\! \mathrm{vec}(\psi(\vect s_k,\vect a_k)\psi(\vect s_k,\vect a_k)\!^\top),
\end{equation*}
where $\mathrm{vec}(\cdot)$ stacks the columns of a matrix into a single
column vector. Then~\eqref{eq:LS:W} is written as $\mathbb{E}[(\hat \psi(\vect s_k,\vect a_k)^\top \mathrm{vec}(W)-\hat \delta_k)^2]$ and similar to~\eqref{eq:ls-g}, the unconstrained optimal solution to~\eqref{eq:LS:W}, denoted by $\hat W$, follows from 
\begin{equation}\label{eq:ls-W}
     \mathbb{E}[\hat \psi(\vect s_k,\vect a_k)\hat \psi(\vect s_k,\vect a_k)^\top]\,  \mathrm{vec}(\hat W)= \mathbb{E}[\hat \delta_k\hat \psi(\vect s_k,\vect a_k)].
\end{equation}
Thus, the optimal solution of the PSD-constrained optimization~\eqref{eq:LS:W}, denoted by $W^\star$, is the Euclidean projection of \(\hat W\) onto the PSD cone, i.e., 
    \begin{equation}\label{eq:thm-proj}
        W^\star
        = \arg\min_{W\succeq 0} \|W - \hat W\|_F^2
        = \Pi_{\mathbb{S}_+^d}(\hat W),
    \end{equation}
    where \(\Pi_{\mathbb{S}_+^d}\) denotes the Frobenius-norm projection onto the PSD cone, i.e., if \(\hat W = U\Lambda U^\top\) is an eigendecomposition, then $W^\star = U\,\max(\Lambda,0)\,U^\top$. The feasible set \(\{W\succeq 0\}\) in ~\eqref{eq:LS:W} is a closed convex cone and the cost is a convex quadratic, the constrained optimum equals the Euclidean projection of the unconstrained minimizer onto the feasible set~\cite{boyd2004convex}.

Finally, given the critic parameters $(\vect v, \vect g, W)$, the actor parameters are updated using the quasi-Newton rule in~\eqref{eq:QN:update}.

The sequence \eqref{eq:lstdv}–\eqref{eq:ls-g}–\eqref{eq:ls-W} implements decoupled matching.  First,  we evaluate the baseline value function, then the linear advantage is fitted which gives the gradient, and finally the quadratic advantage in a lifted space is calculated for the curvature estimation. This decoupling reduces leakage of value–approximation errors into the curvature estimate, yielding a more robust second–order actor step. Note that a similar principle has been observed in nonconvex optimization, where decoupling gradient and curvature updates leads to improved numerical robustness and reduced error interaction between the two estimation stages~\cite{gratton2020decoupled}. Moreover, this sequential estimation of $\vect g$ and $W$ is consistent with Theorem~\ref{thm:comp:2nd}, which shows that the gradient and curvature components of the critic can be identified independently. Hence, the LSTD-based procedure naturally aligns with the theoretical quasi-Newton compatibility conditions.
Expectations are computed as discounted sum over the trajectories and  possibly across multiple episodes. In practice, if any matrix inverse required in the algorithm is numerically ill-conditioned or undefined, it should be replaced by its Moore–Penrose pseudoinverse to ensure numerical stability. A compatible deterministic second-order actor--critic based on batch LSTD algorithm can be summarized in Algorithm~\ref{alg:batch-lstd-qn-actor}.

\begin{algorithm}[!htbp]
\caption{Episodic LSTD algorithm for Quasi-Newton Actor--Critic}
\label{alg:batch-lstd-qn-actor}
\DontPrintSemicolon
\SetKwInOut{Input}{Input}\SetKwInOut{Output}{Output}
\Input{Per-episode horizon $K$; batch size $E$; discount $\gamma$; differentiable policy $\vect\pi_{\vect\theta}(\vect s)$; $\nabla_{\vect\theta}\vect\pi(\vect s)$; exploration scale $\sigma$; step-size $\alpha_\theta$;
features $\phi(\vect s)$, $\psi(\vect s,\!\vect a)\!\!:=\!\!\nabla_{\vect \theta} \vect \pi_{\vect \theta}(\vect s)(\vect a\!-\!\vect\pi_{\vect\theta}(\vect s)\!)$ 
}
\Output{Optimal policy parameters $\vect\theta^\star$.}

\textbf{Initialize} $i\gets 0,\quad$ $\vect\theta_0$.

\While{$\vect\theta$ has not converged}{
  $A_v\gets 0, b_v\gets 0, A_g\gets 0, b_g\gets 0$,\; $A_W\gets 0, b_W\gets 0,\nabla J\gets 0,H\gets 0$.\;
    \tcp{1) System Transition}
  \For{$e=1,\dots,E$}{
    Sample $\vect s^e_1\sim p_1$.\;
    \For{$k=1,\dots,K$}{
      $\vect a^e_k \gets \vect \pi_{\vect\theta_i}(\vect s^e_k) + \vect \epsilon_k$, \quad $\vect \epsilon_k\sim\mathcal{N}(0,\sigma^2I)$.\;
      $\vect s^e_{k+1}\sim p(\cdot\mid \vect s^e_k,\vect a^e_k)$.
    }}
  
  \tcp{2) LS for baseline as in~\eqref{eq:lstdv}}
  \For{$e=1,\dots,E$}{
  \For{$k=1,\dots,K$}{
$A_v \gets A_v +\gamma^{k-1} \phi(\vect s^e_k)\bigl(\phi(\vect s^e_k)-\gamma\phi(\vect s^e_{k+1})\bigr)^\top$.\;
  $b_v \gets b_v+\gamma^{k-1}\ell(\vect s^e_k,\vect a^e_k)\phi(\vect s^e_k)$.
  }}
  $A_v \gets \frac{1}{E}A_v, \quad b_v \gets \frac{1}{E}b_v,\quad \vect v \gets A_v^{-1} b_v$.\;
  
  \tcp{3) LS for gradient as in~\eqref{eq:ls-g}}
 \For{$e=1,\dots,E$}{
  \For{$k=1,\dots,K$}{$\delta^e_k\gets \ell(\vect s^e_k,\vect a^e_k)+\gamma\vect v^\top\phi(\vect s^e_{k+1})-\vect v^\top\phi(\vect s^e_k)$.\; $\psi^e_k\gets \psi(\vect s^e_k,\vect a^e_k)$.\;
  $A_g \gets A_g+\gamma^{k-1}\psi^e_k(\psi^e_k)^\top$. \;
   $b_g \gets b_g+\gamma^{k-1}\delta^e_k\psi^e_k$.
  }}
  $A_g \gets \frac{1}{E}A_g, \quad b_g \gets \frac{1}{E} b_g,\quad\vect g \gets A_g^{-1} b_g$.\;
  
  \tcp{4) LS for Hessian as in~\eqref{eq:ls-W}}
  \For{$e=1,\dots,E$}{
  \For{$k=1,\dots,K$}{ $\hat\delta^e_k\gets\delta^e_k-(\psi^e_k)^\top\vect g$,  $\quad \hat\psi^e_k\gets\mathrm{vec}(\psi^e_k(\psi^e_k)^\top)$.\;
    $A_W \gets A_W+ \gamma^{k-1} \hat\psi^e_k(\hat\psi^e_k)^\top$.\;
  $b_W \gets b_W+ \gamma^{k-1}\hat\delta^e_k\hat\psi^e_k$.
  }}
  $A_W \gets \frac{1}{E}A_W, \quad b_W \gets \frac{1}{E} b_W$.\;
   $\mathrm{vec}(W)\gets A^{-1}_W b_W$,\quad $W\gets\mathrm{reshape}(\mathrm{vec}(W))$.\;
  $W \gets$ {$\Pi_{\mathbb{S}_+^d}(W)$} PSD projection.\; 
  \tcp{\small{5) Actor ingredients as Theorem~\ref{thm:comp:2nd}}}
   \For{$e=1,\dots,E$}{
  \For{$k=1,\dots,K$}{ 
  $\nabla J\gets\nabla J+ \gamma^{k-1} \nabla_{\vect\theta}\vect\pi_{\vect\theta}(\vect s^e_k)\nabla_{\vect\theta}\vect\pi_{\vect\theta}(\vect s^e_k)^\top\vect g$,
  $
  H \gets H+ \gamma^{k-1} \nabla_{\vect\theta}\vect\pi_{\vect\theta}(\vect s^e_k)...$\;$\bigl(\nabla_{\vect\theta}\vect\pi_{\vect\theta}(\vect s^e_k)^\top W \nabla_{\vect\theta}\vect\pi_{\vect\theta}(\vect s^e_k)\bigr) \nabla_{\vect\theta}\vect\pi_{\vect\theta}(\vect s^e_k)^\top.$
  }}
  $\nabla J \gets \frac{1}{E} \nabla J, \quad H \gets \frac{1}{E} H$.
  \;  
  \tcp{\small{6) Quasi-Newton update as in~\eqref{eq:QN:update}}}
  $\vect\theta_{i+1} \gets \vect\theta_i - \alpha_\theta H^{-1}\nabla J$, $\,\,\quad i\gets i+1$
}
\end{algorithm}

The quasi-Newton actor--critic framework developed above can be applied to any differentiable deterministic policy class. In general, the choice of policy parameterization significantly influences the learning performance and stability of actor--critic methods. Various parameterizations, such as neural network policies, affine feedback laws, or optimization-based policies, can be used within this framework. 

\section{Simulation Results}\label{sec:Sim}
In this section, we provide two benchmarks illustrating the efficiency of the proposed approach.
\subsection{Linear Quadratic Regulator (LQR)} 
In the first example, we consider a discounted, linear quadratic regulation (LQR) task with additive noise.
The stochastic dynamics are
$
  \vect s_{k+1} = A\,\vect s_k + B\,\vect a_k + \vect w_k,
$
where
\begin{equation*}
  A=\begin{bmatrix}
  0.95 & 0.20 & 0 \\
  -0.10 & 1.20 & 0.30 \\
  0 & -0.10 & 1.10
  \end{bmatrix},\qquad
  B=\begin{bmatrix}
  0.20 & 0.50\\
  0.10 & -0.50\\
  -0.30 & -0.60
  \end{bmatrix}.
\end{equation*}
The stage cost is quadratic,
$  \ell(\vect s,\vect a) = \|\vect s\|^2+ 10\, \|\vect a\|^2,$ and the discount factor is $\gamma=0.999$.
We use a linear deterministic policy $\vect \pi_{\vect \theta}(\vect s)= -K(\vect \theta)\vect s$ with parameters $\vect \theta=\mathrm{vec}(K)\in\mathbb{R}^{mn}$. The initial state  $\vect s_0\sim \mathcal{N}([5, 5, 5]^\top, 10^{-2} I_{3})$ is drawn from a Gaussian distribution.
The process noise is $\vect w_k\sim\mathcal{N}([0, 0, 0]^\top,10^{-6}I_3)$.
We use the quadratic compatible critic from Section~\ref{sec:QN_AC} with baseline features
\begin{equation*}
\phi(\vect s) = \Big[\,1,\; s_1,s_2,s_3,\; s_1^2,s_2^2,s_3^2,\; s_1s_2,\; s_1s_3,\; s_2s_3\,\Big]^\top \in \mathbb{R}^{10}.
\end{equation*}
We perform $60$ policy updates. Each update uses a batch of $E=500$ independent trajectories of horizon $K=50$ as episodic rollouts. We initialize the policy with
\[
K_0 \;=\;
\begin{bmatrix}
0.1 & 0.1 & 0.1 \\
-0.5 & -0.2 & -0.5
\end{bmatrix},
\quad \vect \theta_0=\mathrm{vec}(K_0).
\]

\begin{figure}[t!]
   \centering
    \includegraphics[width=0.41\textwidth]{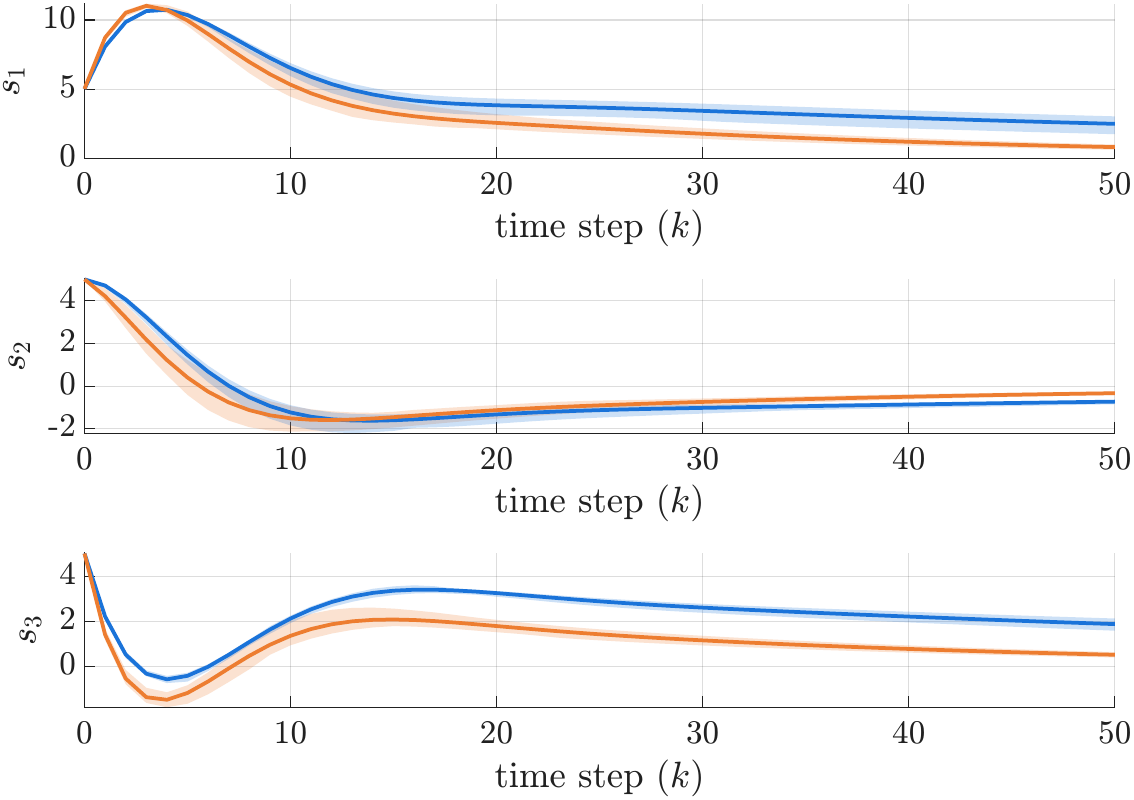}
    \caption{State evolution in the first step (blue) vs.\ the last step (orange) under the quasi-Newton actor--critic.}   \label{fig:states_early_last}
\end{figure}

\begin{figure}[t!]
   \centering
    \includegraphics[width=0.41\textwidth]{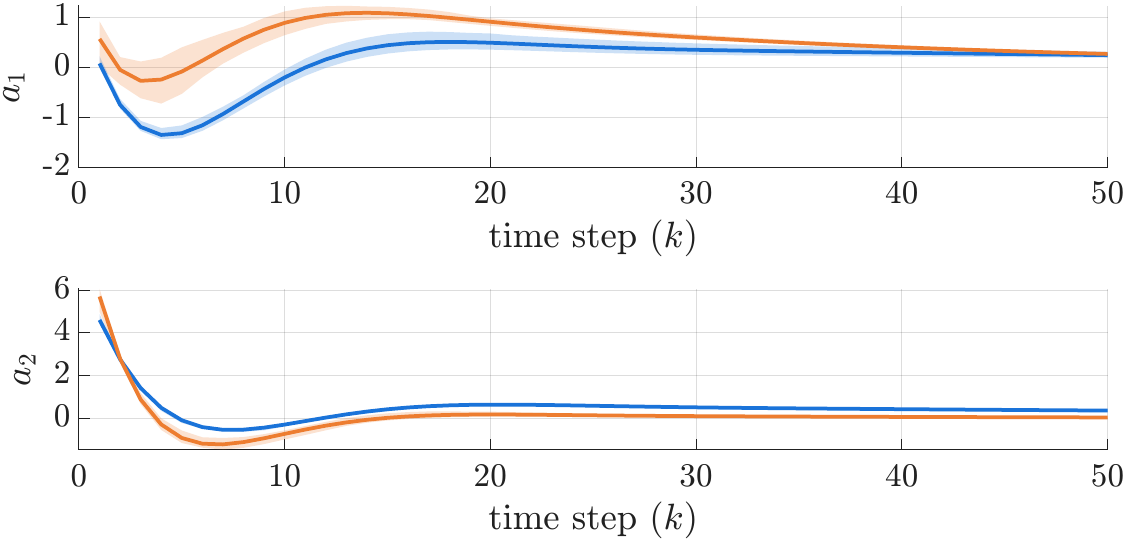}
    \caption{Control actions in the first step (blue) vs.\ the last step (orange) under the quasi-Newton actor--critic.}  \label{fig:actions_early_last}
\end{figure}

\begin{figure}[t!]
   \centering
    \includegraphics[width=0.41\textwidth]{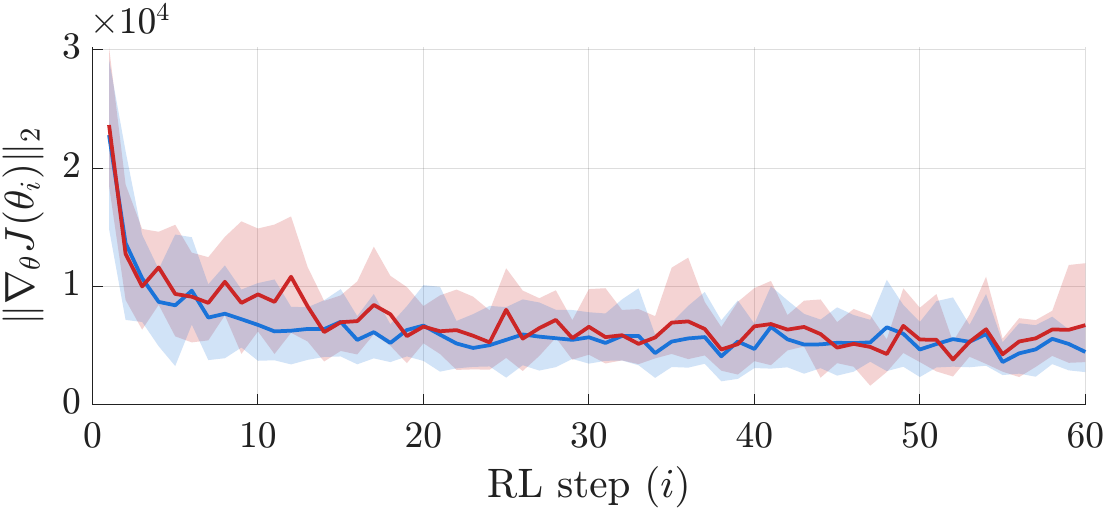}
    \caption{Norm of the performance gradient across policy updates: quasi-Newton (blue) vs.\ first-order DPG (red). The two methods observe comparable gradient magnitudes.}
    \label{fig:grad_norm}
\end{figure}

Figures~\ref{fig:states_early_last} and \ref{fig:actions_early_last} compare the closed-loop trajectories (state and action) from the first training step (blue) and the last step (orange) under the quasi-Newton policy. Faster decay of the state and reduced control effort in the last step indicate a lower accumulated cost. Figure~\ref{fig:grad_norm} shows the norm of the performance gradient versus policy-update index for quasi-Newton (blue) and first-order deterministic policy gradient (red). Both methods track similar gradient magnitudes across updates, confirming that the gradient component of the compatible critic is unbiased and illustrates the role of the Hessian in the RL steps.

\begin{figure}[t!]
   \centering
\includegraphics[width=0.41\textwidth]{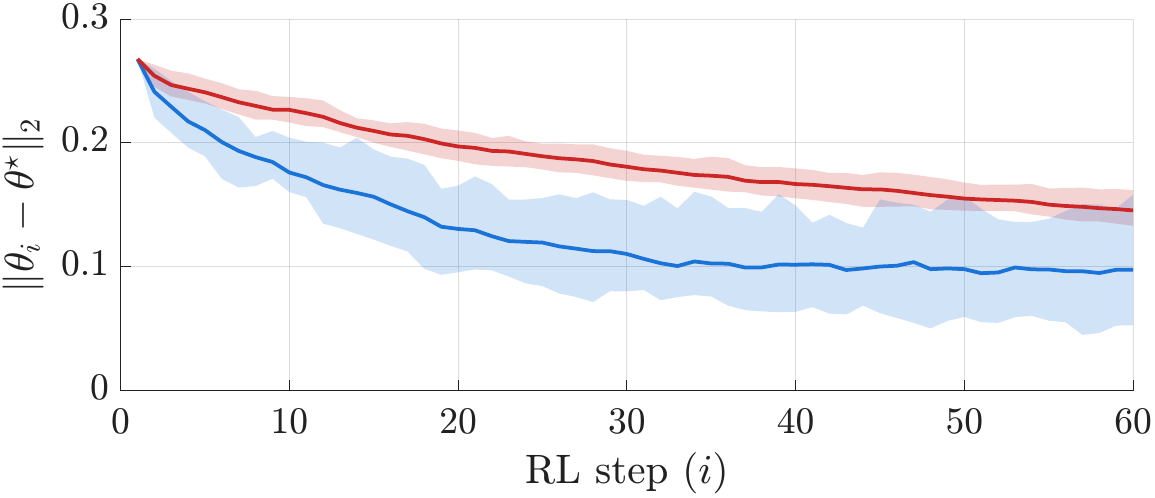}
    \caption{Parameters over updates: quasi-Newton (blue) vs.\ first-order DPG (red). Curvature information yields faster convergence.}
   \label{fig:theta_dist}
\end{figure}

\begin{figure}[t!]
   \centering
    \includegraphics[width=0.41\textwidth]{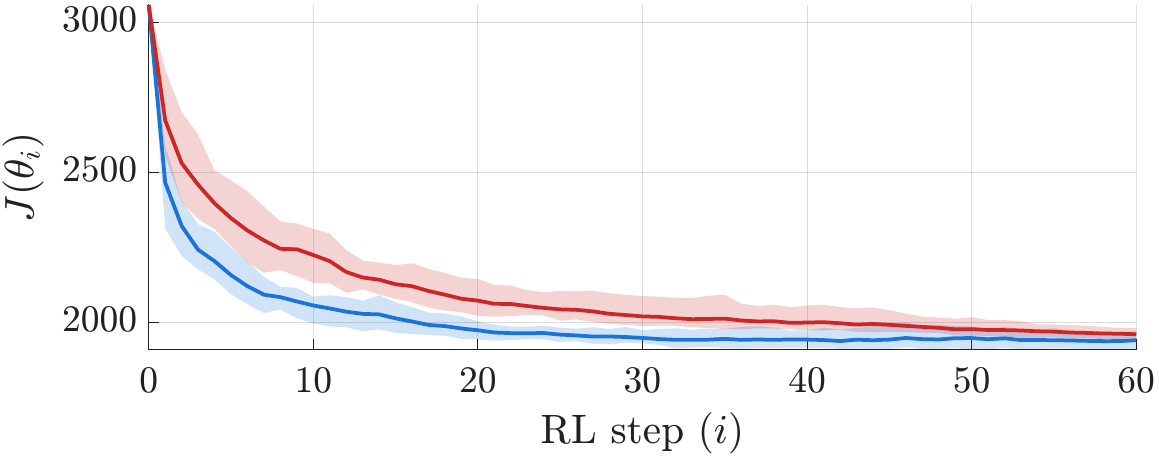}
    \caption{Performance $J(\vect \theta)$ vs.\ policy-update index: quasi-Newton (blue) achieves a faster decrease and reaches a lower value earlier than first-order DPG (red).}
\label{fig:J_vs_update}
\end{figure}

Figure~\ref{fig:theta_dist} shows the distance of policy parameters to the optimal values over updates. Quasi-Newton (blue) converges faster than the first-order method (red), highlighting the benefit of curvature in preconditioning the step. The residual error does not vanish entirely due to the stochasticity inherent in the exploration and the dynamics. Moreover, the increased variance in the quasi-Newton results arises from the second-order exploration terms used for curvature estimation. Figure~\ref{fig:J_vs_update} reports $J(\vect \theta)$ versus RL steps. The quasi-Newton curve (blue) decreases faster and reaches a lower cost earlier than the first-order baseline (red), evidencing improved sample efficiency and the rate of convergence.

\subsection{Cart--Pendulum Balancing}
In the second example, we consider a nonlinear cart--pendulum benchmark controlled by a differentiable policy based on model predictive control (MPC), learned with the proposed quasi-Newton compatible actor--critic. The continuous-time dynamics are 
\begin{subequations}\label{eq:cp}
\begin{align}
  (M+m)\ddot{x}+\frac{1}{2}ml\ddot{\phi}\cos\phi &=\frac{1}{2}ml\dot{\phi}^2\sin\phi+u,\\
  \frac{1}{3}ml^2\ddot{\phi}+\frac{1}{2}ml\ddot{x}\cos\phi &=-\frac{1}{2}mgl\sin\phi,
\end{align}
\end{subequations}
where $M$ and $m$ are the cart mass and pendulum mass, respectively, $l$ is the pendulum length and $\phi$ is its angle from the vertical axis. Force $u$ is the control action, $x$ is the cart displacement and $g$ is gravity. We use the Runge-Kutta $4^{\mathrm{th}}$-order method to discretize \eqref{eq:cp} with a sampling time $\mathrm{d}t=0.1\mathrm{s}$. Then, by adding a process noise, we cast the dynamics in the form of $\vect s^+=\vect f(\vect s,\vect a)+\vect\xi$, where $\vect s=[\dot x,x,\dot \phi,\phi]^\top$ is the state, $\vect a=u$ is the action, $\vect\xi$ is a Gaussian noise and $\vect f$ is a nonlinear function representing \eqref{eq:cp} in discrete time. The aim is to stabilize the system at the origin while respecting a soft constraint $\dot x\geq 0$ that reflects the direction-of-motion preference. Thus, the stage cost is
\begin{equation*}
      \ell(\vect s,\vect a)= \vect s^\top\, \vect s +0.01 \vect a^\top \, \vect a + 100\max\{-\dot x,0\},
\end{equation*}
with discount factor $\gamma=0.95$. In this example, we use MPC as function approximator for generating the parameterized policy based on the deterministic model.  

MPC formulates the control problem as an online optimization over finite-horizon cost that explicitly incorporates the system dynamics, performance objectives, and operational constraints. In~\cite{gros2019data, kordabad2023reinforcement}, it has been formally justified that MPC is able to generate an optimization policy for a given MDP even when an inaccurate model of the system is used in the optimization by tuning MPC cost, constraints, etc. Moreover, compared to neural network policies, MPC requires significantly less training data to produce a reasonable control strategy, since its structure is grounded in known system dynamics and constraint formulations~\cite{seel2022convex}.

At each time, the policy applies the first control of the solution of a finite-horizon MPC with horizon $N=20$ that penalizes quadratic stage and terminal terms together with soft state-inequality slacks. The MPC stage cost is $L_{\vect\theta}(\vect s,\vect a)=\vect s^\top Q_{\vect \theta}^\top Q_{\vect \theta} \vect s+\vect a^\top R_{\vect \theta}^\top R_{\vect \theta} \vect a$ with parameters $Q_{\vect \theta}$ and $R_{\vect \theta}$. The soft state constraint $-\dot x\!\le \beta+\sigma$ is used in the MPC scheme with scalar threshold parameter $\beta$ and slack $\sigma\ge 0$. The equality constraints use the \emph{deterministic} nonlinear model and feasibility is enforced via slacks with large penalties. The policy sensitivity $\nabla_{\vect\theta}\vect\pi_{\vect \theta}(\vect s)$ is obtained by implicit differentiation of the Karush–Kuhn–Tucker (KKT) conditions associated with the MPC scheme.

\begin{figure}[t!]
   \centering
    \includegraphics[width=0.41\textwidth]{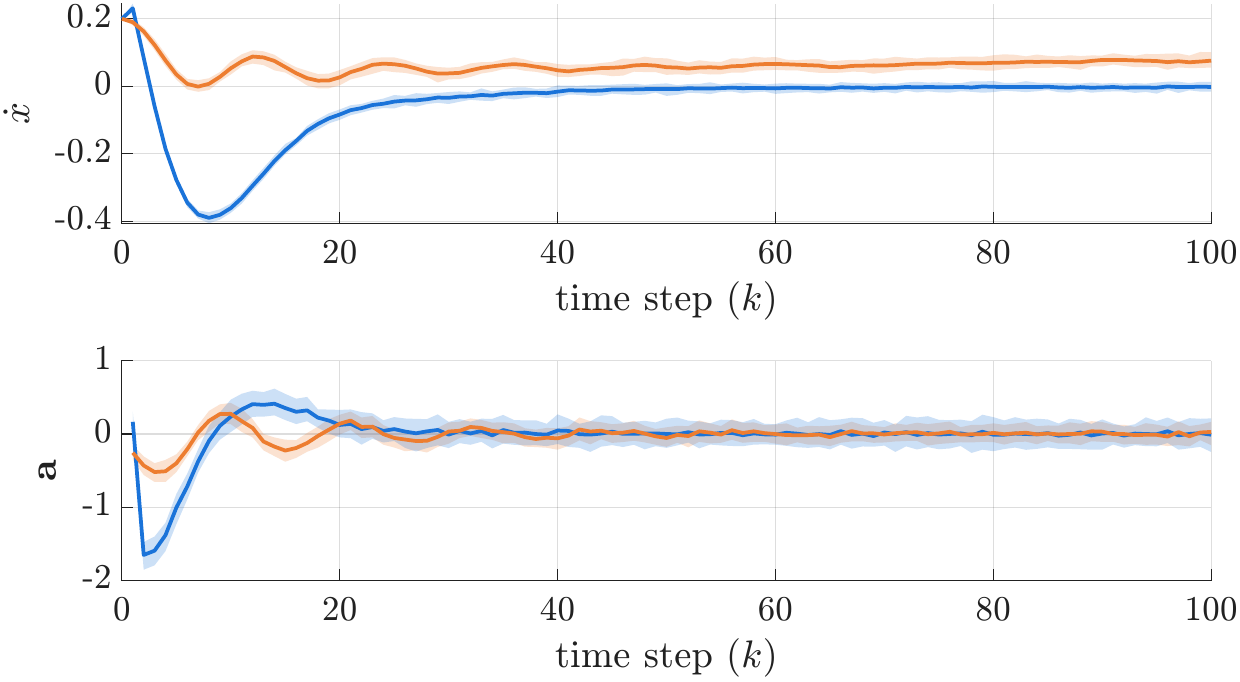}
    \caption{Closed-loop evolution of the first state $\dot x$ and control action under the quasi-Newton actor--critic policy. Blue: first RL step. Orange: last RL step.}
\label{fig:1}
\end{figure}

We run $50$ policy updates. Each update uses $E=50$ episodes of length $K=100$, with identical initial state $\vect s_0=[0.2,\,0.5,\,0.5,\,0.2]^\top$ and exploration $\vect a_k=\vect \pi_{\vect\theta}(\vect s_k)+\varepsilon_k$, $ \varepsilon_k\sim\mathcal{N}(0,0.01)$. Moreover, the baseline state feature vector $\phi(\vect s)$ is a full quadratic monomial feature set in the four-dimensional state $\vect s$. Figure~\ref{fig:1} shows the closed-loop trajectories of the first state $\dot x$ and the control action for the first RL update (blue) and the last update (orange) obtained using the proposed quasi-Newton actor--critic method. At the beginning of training, the soft constraint $\dot x\geq 0$ is violated more, while the final policy successfully enforces the constraint, keeping the velocity non-negative. The figure also illustrates that the control effort required in the last RL step is smaller, indicating improved performance of the policy.

\begin{figure}[t!]
   \centering
    \includegraphics[width=0.41\textwidth]{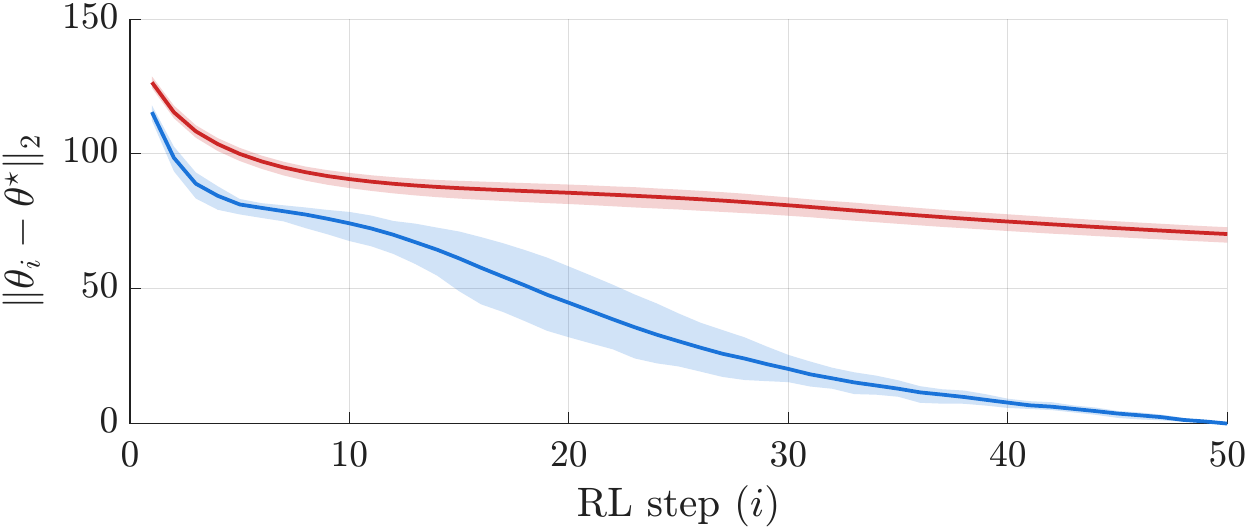}
    \caption{Parameter distance $|\vect{\theta}-\vect{\theta}^\star|_2$ over RL updates for quasi-Newton (blue) and first-order (red) actor--critic methods.}
\label{fig:2}
\end{figure}

\begin{figure}[t!]
   \centering
    \includegraphics[width=0.41\textwidth]{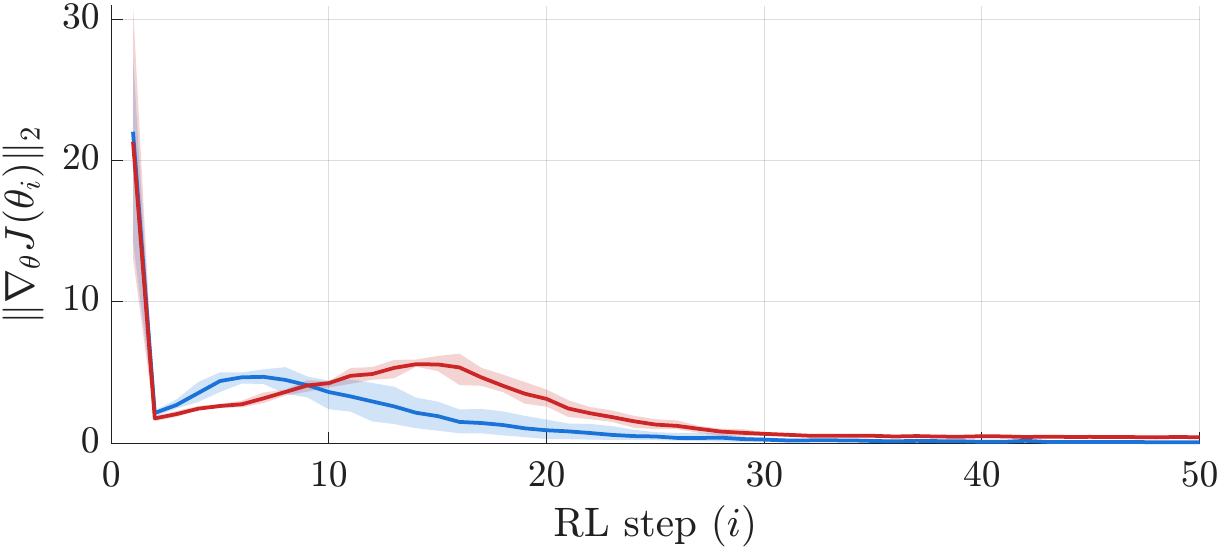}
    \caption{Norm of the policy gradient during learning.}
\label{fig:3}
\end{figure}

\begin{figure}[t!]
   \centering
    \includegraphics[width=0.41\textwidth]{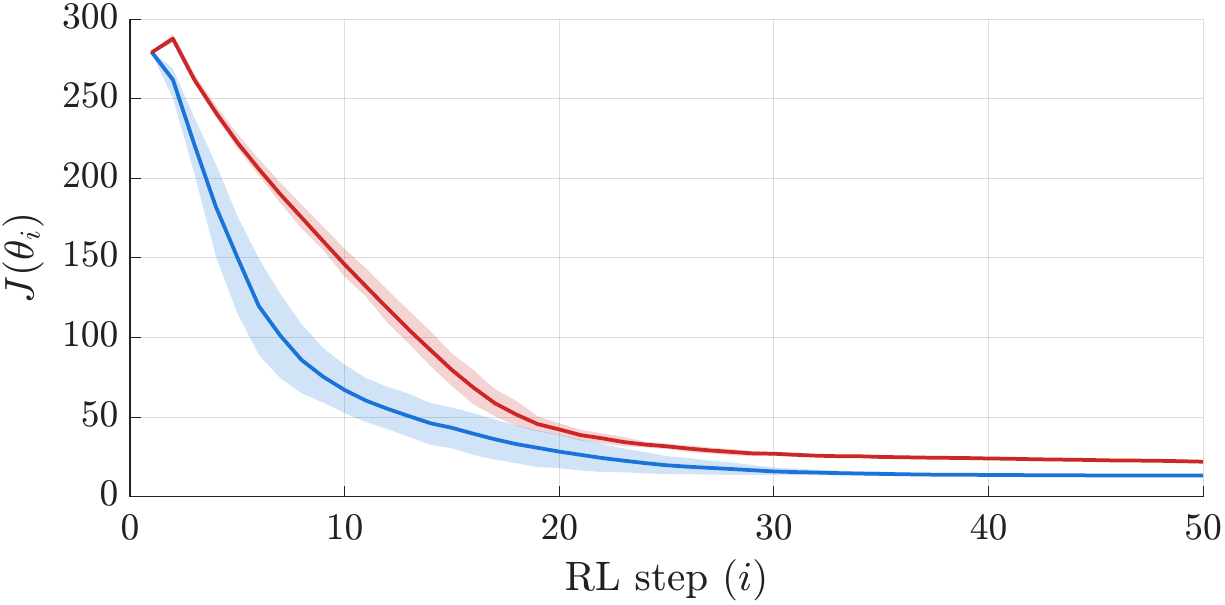}
    \caption{Performance evolution over RL steps for quasi-Newton (blue) and first-order (red) actor--critic methods.}
\label{fig:4}
\end{figure}

Figure~\ref{fig:2} compares the distance of the policy parameters from the final converged value, $\|\vect\theta-\vect\theta^\star\|_2$ for the quasi-Newton (blue) and first-order (red) actor--critic algorithms. The quasi-Newton method converges faster, demonstrating the advantage of incorporating curvature information into the actor update. The policy-gradient norm evolution is shown in Figure~\ref{fig:3}. Both methods successfully drive the gradient toward zero, confirming convergence to a stationary point of the performance function. However, the quasi-Newton actor--critic exhibits a faster gradient decay. Finally, Figure \ref{fig:4} compares the performance $J(\vect\theta)$ across RL steps. The quasi-Newton approach consistently achieves a faster reduction of the performance cost and reaches a lower steady-state value compared with the first-order baseline, confirming its superior sample efficiency and learning speed.

\section{Conclusion}\label{sec:con}
This paper presented a quasi-Newton deterministic actor--critic framework that incorporates second-order curvature information into policy updates to accelerate and stabilize learning. Building on the idea of compatible function approximation, a \emph{quadratic critic} was introduced that preserves both the deterministic policy gradient and the quasi-Newton approximation of the performance Hessian. The resulting structure enables curvature-aware policy updates while maintaining consistency with the underlying deterministic policy gradient theorem. To make the approach practical, a least-squares temporal difference learning scheme was developed for estimating the critic parameters efficiently from batches of data, yielding a sample-efficient and numerically robust algorithm.  The approach can be applied to a wide range of differentiable policy classes, including neural-network and optimization-based policies, making it suitable for control-oriented applications where data efficiency and stability are critical. Simulation results demonstrated that the proposed method achieves faster convergence compared with first-order deterministic actor--critic baselines. Note that although each quasi-Newton update is computationally more expensive, the improved sample efficiency and faster convergence typically offset this cost, leading to superior overall performance per sample and iteration. Future research directions include extending the quasi-Newton actor--critic framework to discrete and hybrid state–action MDPs and reducing variance in Hessian estimation by leveraging smoother action--value approximators, such as model predictive control–based critics or other structured function classes that naturally enforce curvature regularity.
\vspace{-0.4cm}
\bibliographystyle{IEEEtran}
\bibliography{QuasiAC}
\end{document}